\documentclass[10pt]{article}
\usepackage[utf8]{inputenc}
\usepackage[T1]{fontenc}
\usepackage{absspconf,amsmath}
\usepackage{tikz}
\usepackage{amssymb}% http://ctan.org/pkg/amssymb
\usepackage{tabularx}
\usepackage{multirow}
\usepackage{siunitx}
\usepackage{pifont}% http://ctan.org/pkg/pifont
\usepackage{graphicx}
\usepackage{amsfonts}
\usepackage{booktabs}
\usepackage{blindtext}
\usepackage{csquotes}
\usepackage{makecell}
\usepackage{tikz}
\usepackage{tikzscale}
\usepackage{gnuplot-lua-tikz}
\usepackage{microtype}
\usepackage{float}
\usepackage[above]{placeins}

\usepackage{array}
\usepackage[style=ieee, doi=false, url=false, isbn=false, minnames=1, maxnames=3, date=year, hyperref=true]{biblatex}
\defbibheading{secbib}[\bibname]{%
  \section{#1}%
  \markboth{#1}{#1}
}
\renewbibmacro{in:}{}
\DeclareBibliographyDriver{unpublished}{%
    \usebibmacro{bibindex}%
    \usebibmacro{begentry}%
    \usebibmacro{author/translator+others}%
    \newunit
    \usebibmacro{title}%
    \newunit\newblock
    \usebibmacro{eprint}%
    \newunit\newblock
    \usebibmacro{date}
    \usebibmacro{finentry}%
}
\DeclareFieldFormat{eprint:arxiv}{%
  arXiv\addcolon\space
  \ifhyperref
    {\href{https://arxiv.org/abs/#1}{%
       \nolinkurl{#1}%
       \iffieldundef{eprintclass}
         {}
         }}
    {%\nolinkurl{#1}%
    #1
     \iffieldundef{eprintclass}
       {}
       }
}
\usepackage{caption, setspace}
\usepackage{soul}
\usepackage[acronym]{glossaries}
\usepackage{etoolbox}

\usepackage{comment}
\usepackage{todonotes}
\usepackage[hidelinks]{hyperref}
\usepackage[capitalise,noabbrev]{cleveref} % only cleveref should come after hyperref

\newcommand{\ubold}{\fontseries{b}\selectfont}

\newcommand{\y}{\boldsymbol{y}}
\newcommand{\z}{\boldsymbol{z}}
\newcommand{\sctf}{\emph{Synchronized Class Token Fusion} (SCT Fusion)}
\newacronym{msi}{MSI}{multispectral imaging}
\newacronym{dl}{DL}{deep learning}
\newacronym{sar}{SAR}{synthetic aperture radar}
\newacronym{rs}{RS}{remote sensing}
\newacronym{cnn}{CNN}{convolutional neural network}
\newacronym{vit}{ViT}{vision transformer}
\newacronym{lidar}{LIDAR}{light detection and ranging}
\newacronym{hsi}{HSI}{hyperspectral imaging}
\newacronym{mml}{MML}{multi-modal~learning}
\newacronym{lstm}{LSTM}{long short-term memory}

\makeatletter
\let\oldmakefirstuc\makefirstuc
\renewcommand*{\makefirstuc}[1]{%
  \def\gls@add@space{}%
  \mfu@capitalisewords#1 \@nil\mfu@endcap
}
\def\mfu@capitalisewords#1 #2\mfu@endcap{%
  \def\mfu@cap@first{#1}%
  \def\mfu@cap@second{#2}%
  \gls@add@space
  \oldmakefirstuc{#1}%
  \def\gls@add@space{ }%
  \ifx\mfu@cap@second\@nnil
    \let\next@mfu@cap\mfu@noop
  \else
    \let\next@mfu@cap\mfu@capitalisewords
  \fi
  \next@mfu@cap#2\mfu@endcap
}
\makeatother

\title{Transformer-based Multi-Modal Learning for Multi-Label Remote Sensing Image Classification}
\name{David Sebastian Hoffmann$^{*\,1}$, Kai Norman Clasen$^{*\,1}$, Beg\"{u}m Demir$^{1,2}$
 }
\address{
    $^1$Faculty of Electrical Engineering and Computer Science, Technische Universität Berlin, Germany\\%
    $^2$BIFOLD -- Berlin Institute for the Foundations of Learning and Data, Germany
}

\begin{document}

\maketitle
\def\thefootnote{*}\footnotetext{These authors contributed equally to this work}\def\thefootnote{\arabic{footnote}}

\begin{abstract}
In this paper, we introduce a novel \emph{Synchronized Class Token Fusion} (SCT Fusion) architecture in the framework of multi-modal multi-label classification (MLC) of \gls{rs} images. 
The proposed architecture leverages modality-specific attention-based transformer encoders to process varying input modalities, while exchanging information across
modalities by \emph{synchronizing} the special class tokens after each transformer encoder block.
The synchronization involves fusing the class tokens with a trainable fusion transformation, resulting in a synchronized class token that contains information from all modalities.
As the fusion transformation is trainable, it allows to reach an accurate representation of the shared features among different modalities.
Experimental results show the effectiveness of the proposed architecture over single-modality architectures and an early fusion multi-modal architecture when evaluated on a multi-modal MLC dataset. 
The code of the proposed architecture is publicly available at \url{https://git.tu-berlin.de/rsim/sct-fusion}.
\glsresetall
\end{abstract}

\begin{keywords}
Multi-modal fusion, multi-label image classification, deep learning, transformer, remote sensing.
\end{keywords}

\begin{figure*}
    \centering
    \includegraphics[width=.725\linewidth]{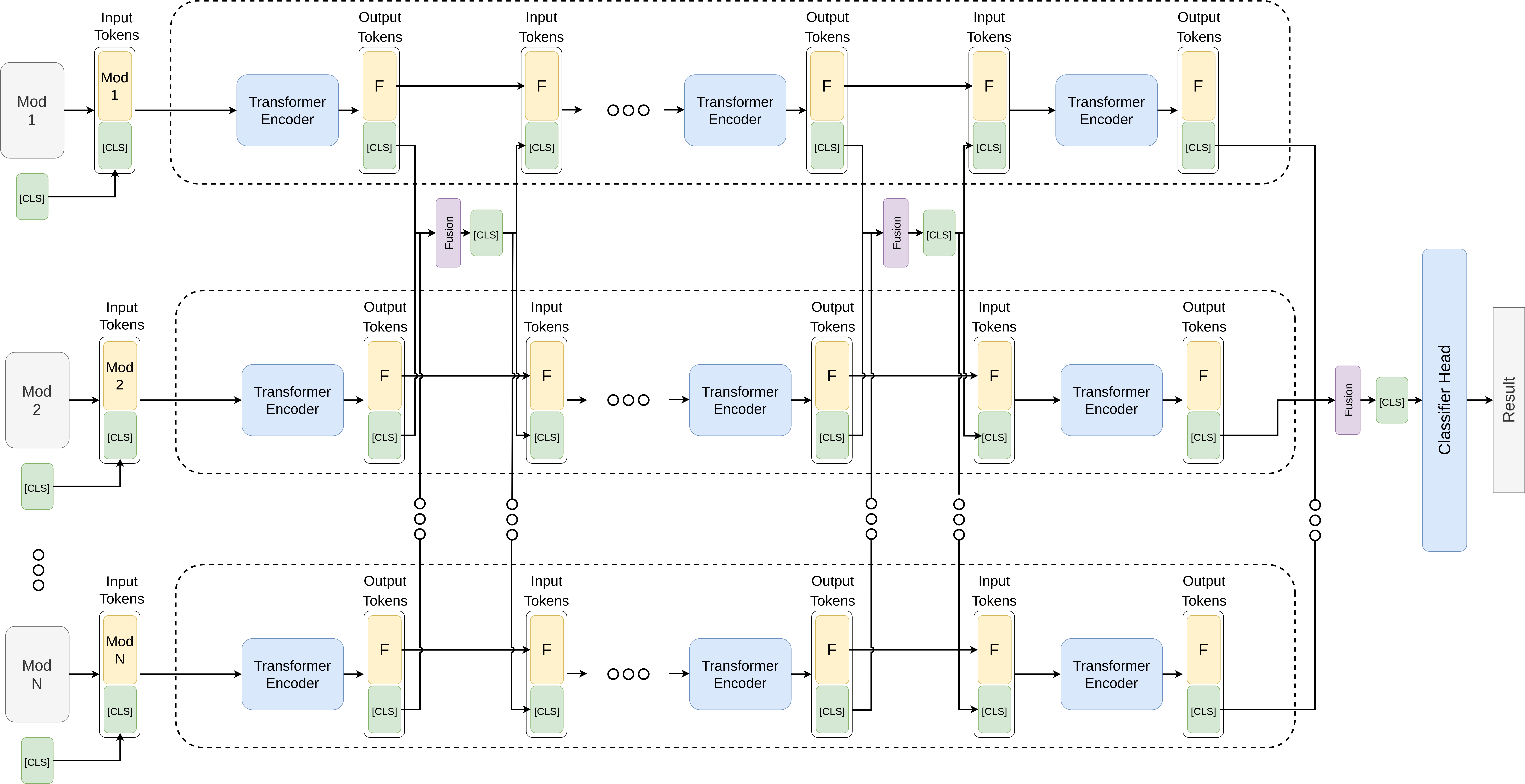}
    \caption{\label{fig:sct_fusion} Illustration of the full architecture design for the proposed SCT Fusion architecture for $N$ input modalities, where \texttt{[CLS]} and \texttt{F} refer to the class and feature tokens, respectively.
    }
    \vspace{-1em}
\end{figure*}

\section{Introduction}
The ever-growing archives of \gls{rs} images are a valuable source of information for monitoring the Earth's surface.
One of the most important learning tasks is multi-label classification, which aims at automatically assigning multiple land-cover class labels (i.e., multi-labels) to each \gls{rs} image scene in an archive.
In recent years, \gls{dl} techniques have attracted substantial attention in the field of multi-label classification of \gls{rs} images.
This is largely due to their high capability to capture the complex spatial and spectral characteristics of these images.
Numerous architectures have been proposed, including specialized \glspl{cnn} \cite{mlcdl}, hybrid architectures incorporating \glspl{cnn} and \gls{lstm} components \cite{sumbul2020}, and recently, transformer-based architectures \cite{mlctransformer}.
The latter, in particular, are rising in popularity due to the benefits of self-attention, which is a fundamental operation in transformers.
Self-attention capitalizes on non-local relationships within an image, enabling the learning of long-range contextual and spatial associations among various image components while mitigating the intrinsic inductive bias of CNNs.
These properties have demonstrated their effectiveness in tackling multi-label classification problems \cite{mlctransformer}.
However, the pursuit of accurate multi-label scene classification is not necessarily limited to a single modality.
Recent studies have shown that multi-modal images associated with the same geographical area allow for a rich characterization of RS images when jointly exploited and thus improve the effectiveness of the selected image analysis task \cite{fusion_comparison}.
Accordingly, to jointly exploit multi-modal RS images in the framework of \gls{mml}, the development of multi-modal RS image classification methods has attracted great attention in RS.
Notably, the aforementioned transformer-based architectures are becoming a popular choice in the MML research \cite{multimodal_transformer_survey}.
Despite the extensive development of multi-modal transformer-based architectures for pixel-based classification tasks, their application in multi-modal multi-label RS image classification remains relatively unexplored \cite{deep_hierarchical_vit}.

In this paper, we propose a novel transformer-based multi-modal learning architecture tailored for multi-modal multi-label classification problems in RS, illustrated in \cref{fig:sct_fusion}.
Specifically, we introduce \emph{Synchronized Class Token Fusion} (SCT Fusion), which is an architecture derived from the popular \gls{vit} \cite{vit} architecture.
The standard \gls{vit} architecture operates as follows: 1) the input image is split into non-overlapping patches; 2) the patches are transformed into patch embedding tokens; 3) an additional learnable class token is appended to the patch embedding token sequence; 4) the tokens are processed by several transformer encoder blocks \cite{vit} performing consecutive self-attention computations among the patch embedding tokens and the class token; and 5) finally, the class token of the last layer's output sequence is extracted and used for the classification step.
Our proposed architecture employs a dedicated ViT for each of the input modalities and applies a repeated \emph{synchronization} of the class tokens,
where the class tokens of all modalities are fused and the result redistributed after every transformer encoder block, thereby introducing information from all other modalities to each modality-specific ViT.
By accurately incorporating inputs from different modalities, the proposed architecture leads to a superior multi-modal data representation and improves classification performance.
To the best of our knowledge, a transformer-based multi-modal fusion approach relying on a repeated synchronization of class tokens has not been investigated in the RS domain.

\section{Proposed Transformer-based Multi-Modal Fusion Method}

Let $\big\{\mathcal{X}^1, \ldots, \mathcal{X}^N\big\}$  be a set of $N$ \gls{rs} image archives where each archive corresponds to a unique image modality captured by a distinct sensor.
Each archive $\mathcal{X}^j$ consists of $M$ images and is defined as $\mathcal{X}^j = \big\{\boldsymbol{x}_i^j\big\}_{i=1}^M$, where $\boldsymbol{x}_i^j$ is the $i$-th RS image in the $j$-th image archive. 
Let $\big\{\boldsymbol{x}_i^j\big\}_{j=1}^N$ be the $i$-th multi-modal image set that includes $N$ images acquired by different sensors on the same geographical area.
Each image is defined by $\boldsymbol{x}_i^j \in \mathbb{R}^{h_j \times w_j \times c_j}$, where $h_j, w_j, c_j \in \mathbb{N}$ refer to the height, width and channel dimensions, respectively.
Each image set $\big\{\boldsymbol{x}_i^j\big\}_{j=1}^N$ is annotated by one or more labels from a set of class labels $\{L_1, \dots, L_l\}$ with $l$ denoting the number of labels in the archive.
This is represented by a multi-hot-encoded vector $\boldsymbol{y}_i \in \{0, 1\}^l$, where each entry corresponds to the presence or absence of class $L_j$ in the geographical area covered by $\big\{\boldsymbol{x}_i^j\big\}_{j=1}^N$.
The task of multi-label classification can therefore be defined as finding a function $f$ that takes $\big\{\boldsymbol{x}_i^j\big\}_{j=1}^N$ as input to generate a vector $\hat{\y}_i \in {[0,1]}^l$ so that $\hat{\y}_i$ most closely approximates $\y_i$ under a chosen loss function.

To achieve accurate multi-modal multi-label classification, we propose the \sctf{} architecture.
The proposed multi-modal architecture aims at learning a joint feature representation from different image modalities by applying a repeated synchronization of class tokens to consistently fuse the deep features of the modalities while taking advantage of the special class token.
To this end, our SCT Fusion method utilizes a dedicated \gls{vit} for each input modality, where the \gls{vit} architecture operates on a sequence of feature tokens derived from an image of each modality by a specialized embedding layer. 
Each image $\boldsymbol{x}^j$ is divided into multiple patches $\boldsymbol{x}_{p_j}^j \in \mathbb{R}^{p_j \times p_j \times c_j}$ with a spatial dimension of $p_j \in \mathbb{N}$ so that $h_j/p_j \in \mathbb{N}$ and $w_j/p_j \in \mathbb{N}$. 
A learnable linear transformation is then applied to each $\boldsymbol{x}_{p_j}^j$ to derive a patch embedding token $\z^j \in \mathbb{R}^{d_e}$, where $d_e \in \mathbb{N}$ denotes the embedding dimension.
These tokens condense the information present in the pixels of $\boldsymbol{x}_{p_j}^j$ into a lower dimensional feature space.
In addition to the patch tokens, a special learnable class token $\z_{cls}^j \in \mathbb{R}^{d_e}$ is appended to the token sequence.
Note that we require each token $\z^j \in \mathbb{R}^{d_e}$ to share the same dimensionality for all image modalities $j \in \{1, \ldots{}, N\}$. 
The entire token sequence derived from $\boldsymbol{x}^j$ can hence be defined as $\mathcal{Z}^j = \{\z_1^j, \z_2^j, \cdots , \z_{k_p}^j, \z_{cls}^j\}$, where $k_p \in \mathbb{N}$ shows the number of patches. 
The input sequence $\mathcal{Z}^j$ is then processed by a modality-specific transformer encoder block, where the self-attention operation is applied.
For a detailed definition of the transformer encoder block, the reader is referred to \cite{vit}.
The output of the modality-specific transformer encoder is forwarded to the next transformer encoder block, except for the class token, which is first synchronized across all modalities.
The synchronization of the class tokens is realized by concatenating the class tokens along the feature dimension to form a combined class token representation $\z_{cls}^{\text{concat}} \in \mathbb{R}^{N d_{e}}$ and fusing them to a synchronized class token $\z_{cls}^*$ by applying a trainable fusion transformation $g: \mathbb{R}^{N d_{e}} \rightarrow \mathbb{R}^{d_{e}}$.
One such learnable fusion transformation may be defined as a linear layer:
\begin{equation}
    \z_{cls}^* = \boldsymbol{W}^{cls} \z_{cls}^\text{concat} + \boldsymbol{b}^{cls},
    \label{equ:cls_token_fusion}
\end{equation}
where $\boldsymbol{W}^{cls} \in \mathbb{R}^{{d_{e} \times (N d_{e})}}$ and $\boldsymbol{b}^{cls} \in \mathbb{R}^{d_{e}}$ refer to the learned weights and bias parameters of the linear layer, respectively.
As a trainable transformation performs the fusion step, the weights can adapt for each input modality individually.
Additionally, the projection of the class tokens $\z_{cls}^\text{concat}$ into a lower dimensional feature space $\mathbb{R}^{d_{e}}$ should encourage the selection of features with higher importance for the classification task \cite{attnbottlenecks}.
Note that compared to the bottleneck tokens introduced in \cite{attnbottlenecks} that are attached to the token sequences and passively passed through the encoders, our architecture actively applies a configurable fusion transformation to the existing classification tokens.
After fusing the class tokens, the synchronized class token $\z_{cls}^*$ is passed back to each modality-specific encoder for further processing instead of the modality-specific class token $\z_{cls}^{j}$ of the previous token sequence. % blocks' output
The process of passing the full token sequences through transformer encoder blocks and synchronizing the class tokens is repeated $r \in \mathbb{N}$ times for each modality-specific encoder.
As a result of the synchronization of the class tokens, information from the other modalities is introduced in the modality-specific encoders.
Finally, the last synchronized class token $\z_{cls}^*$ is passed to a classification layer to compute the class prediction scores.
After an end-to-end training of the full network, the proposed architecture is applied to each image in the archives and assigns multi-labels to it.

\section{Experimental Results}
We evaluate the multi-label classification performance of our proposed SCT Fusion architecture on the BigEarthNet-MM dataset \cite{Sumbul2021}, which comprises 590,326 pairs of Sentinel-1 \Gls{sar} and Sentinel-2 multispectral images.
Each pair was acquired on the same geographical area over a similar period.
We use the recommended 19 class nomenclature and original dataset split as defined in \cite{Sumbul2021} to ensure the comparability of the results.
The lower-resolution \SI{20}{\metre} bands of the Sentinel-2 images are up-scaled via bilinear interpolation, while the \SI{60}{\metre} bands of the Sentinel-2 images were excluded from the experiments due to their negligible impact on the overall performance but higher processing requirements.
The results that were obtained by training a standard \Gls{vit} model by using either Sentinel-1 images (denoted as Sentinel-1 ViT) or Sentinel-2 images (denoted as Sentinel-2 ViT) are provided as reference baselines.
We further compare our proposed SCT Fusion architecture with an \emph{early fusion} \cite{fusion_comparison} approach, where the input modalities are stacked along their channel dimension and passed through the standard \Gls{vit} architecture without any modifications.
All models utilized a \Gls{vit}-base model with a patch size of $15$, a depth of $8$, and an embedding dimension of $256$ with $8$ attention heads.
For training, we employed the Adam optimizer and trained for $60$ epochs with an initial learning rate of $10^{-4}$.
Note that the proposed SCT Fusion architecture does not require the patch size or the spatial dimensions to be equally sized for each input modality but it was a design choice for the experimental setup.
The multi-modal fusion models were trained with the following augmentations: i) RandomSensorDrop \cite{ben_mm_transformer}; ii) flipping; and iii) cropping, where flipping and cropping were  applied to each modality independently.
Additionally, all models were trained with a stochastic depth regularisation \cite{stochastic_depth} of 0.25.
As the fusion transformation for the SCT Fusion architecture, we have selected a linear layer as described in equation~1. % hard-coded to adjust for others taset % \cref{equ:cls_token_fusion}.
Two average precision (micro and macro averaged) metrics, denoted as AP~(micro/macro), are provided to quantify the performance as well as the macro-averaged $F_2$ score.
\Cref{tab:fusion_performance_msi_sar} shows the results of our experiments.
By assessing the table, one can observe that the proposed architecture leads to the highest scores under all metrics.
Specifically, our proposed architecture achieves an AP (macro) improvement of more than \SI{15}{\percent} and approx.\ \SI{1.8}{\percent} in comparison to the single modality Sentinel-1 and Sentinel-2 ViT baseline results, respectively.
Compared to the multi-modal early fusion method, the proposed SCT Fusion architecture leads to an improvement of over \SI{1}{\percent} for the AP (macro) metric and $F_2$ score.
These results underscore the effectiveness of the class token synchronization in accurately modeling multi-modal images.
It is worth noting that a CNN-based (ResNet-50) early fusion architecture, as reported in \cite{Sumbul2021}, results in an $F_2$ score of \SI{67.23}{\percent} on the BigEarthNet-MM dataset. 
The substantial difference in performances between the CNN-based and transformer-based architectures indicates that the latter are generally more effective for multi-label classification tasks.

\begin{table}
  \small
  \center
  \sisetup{
    round-mode=places,
    detect-all=true,
    detect-weight=true,
    mode=text,
    % Hack to conver the data to the percentage values without having to rewrite the numbers
    % PS: The hack is to 'hide' the fixed exponent x 10^-2
    fixed-exponent=-2, 
    exponent-mode=fixed,
    table-format = 2.2,
    round-precision=2,
  }
  \caption{The results achieved by the proposed SCT Fusion and the baseline architectures for the classification of the images in the BigEarthNet-MM dataset.}
  \label{tab:fusion_performance_msi_sar}
    \begin{tabular}{m{.3\linewidth} S S S S}
        \toprule
        \multicolumn{1}{c}{Model Name} & {{AP} (micro)} & {{AP} (macro)} & {$F_2$ (macro)} 
            \\
        \midrule \vspace{0.2em}
        {Sentinel-1 ViT} & 0.8128 & 0.6796 & 0.5581 %F2_mic 0.6725
            \\ \vspace{0.2em}
        {Sentinel-2 ViT} &  0.8910 & 0.8159 & 0.71251 %F2_mic 0.77296
            \\ \vspace{0.2em}
        % Patch size of 15%
        {Early fusion} & 0.8956 & 0.8227 & 0.7253
            \\ \vspace{0.2em}
        {Proposed SCT\newline Fusion} & \ubold 0.9015 & \ubold 0.8343 & \ubold 0.7369 %& \ubold 0.7905
            \\
    \bottomrule
  \end{tabular}
\end{table}

\begin{figure}
    \centering
    \includegraphics[width=\linewidth]{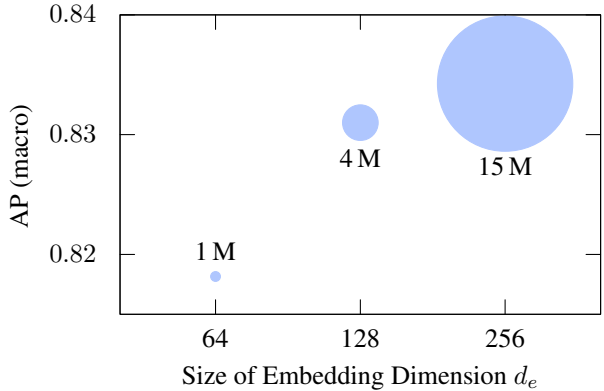}
    \caption{\label{fig:params} 
    % Illustration of the full architecture design for the proposed SCT Fusion architecture for $N$ input modalities, where \texttt{[CLS]} and \texttt{F} refer to the class and feature tokens, respectively.
    The AP~(macro) classification results of the proposed SCT Fusion architecture with different sizes for the embedding dimensions.
    The size of the circle indicates the total number of parameters for the specific architecture.
    }
    \vspace{-1em}
\end{figure}

A notable disadvantage of the proposed architecture is the increased computational complexity in comparison to single-modality and early-fusion architectures, as each image modality necessitates its own dedicated encoder.
However, by tuning several hyper-parameters in the SCT Fusion architecture, such as depth $r$, embedding size $d_e$, number of attention heads, and patch size $p_j$ of the transformer encoders the computational demands can be adjusted.
Note that the patch size $p_j$ and the number of attention heads can be chosen independently for each modality.
To investigate the effect of the size of the embedding dimension $d_e$ on the multi-label classification performance, we have visualized
the AP~(macro) performance given a specific embedding dimension while reporting the number of parameters for each architecture
in \cref{fig:params}.
These results show that considerably smaller models that only marginally compromise the multi-label classification performance can be found. 
Consequently, the computational complexity of our proposed architecture can be substantially alleviated.

\section{Conclusion and Discussion}
In this paper, we introduce a novel \emph{Synchronized Class Token Fusion} (SCT Fusion) architecture in the framework of multi-modal multi-label image classification problems in \gls{rs}.
The proposed architecture learns a joint feature representation from multiple image modalities by consistently exchanging information between the deep features of modality-specific transformer encoders.
In contrast to existing architectures, SCT Fusion enhances information sharing across various modalities by \emph{synchronizing} the class tokens of modality-specific encoders. 
This is achieved by fusing the class tokens after each transformer encoder block using a trainable fusion transformation and subsequently feeding the resulting token back into the modality-specific encoders.
Experimental results show that the proposed architecture outperforms the early fusion approach due to the accurate description of the modality-shared features in multi-label classification problems.
As a final remark, we would like to note that the computational complexity of the proposed architecture is high when compared to the early fusion method with the same base transformer
encoder configuration. % with the same amount
However, additional experiments show that by fine-tuning the modality-specific model parameters, the overall size of the proposed architecture can be significantly reduced while only marginally compromising the classification performance. % only a minimal compromise in classification performance.
As future works, we plan to explore alternative lightweight models, more complex fusion transformations, as well as asymmetric configurations of the modality-specific encoders.

\section{Acknowledgement}
This work is supported by the European Research Council (ERC) through the ERC-2017-STG BigEarth Project under Grant 759764 and by the European Space Agency through the DA4DTE (Demonstrator precursor Digital Assistant interface for Digital Twin Earth) project and by the German Ministry for Economic Affairs and Climate Action through the AI-Cube Project under Grant 50EE2012B.

\vspace{-.5em}
\small
\printbibliography[heading=secbib, title=REFERENCES]
\end{document}